\pdfoutput=1
\documentclass[journal]{IEEEtran}

\usepackage{graphicx}
\usepackage{amsmath}
\usepackage{amssymb}
\usepackage{booktabs}
\usepackage{multirow}
\usepackage{cite}
\usepackage{xcolor}
\usepackage{array}

\begin{document}

\title{Interpretable Physics Informed WiFi Indoor Localization: Learning an Effective Access Point Geometry and Using It to Prune}

\author{
\begin{minipage}[t]{0.48\textwidth}
\centering
Arshia Eftekharizadeh\\
School of Electrical and Computer Engineering\\
University of Tehran\\
Tehran, Iran\\
arshiaeftekhari@ut.ac.ir
\end{minipage}%
\begin{minipage}[t]{0.48\textwidth}
\centering
Rezvan Nasiri\\
School of Electrical and Computer Engineering\\
University of Tehran\\
Tehran, Iran\\
rezvan.nasiri@ut.ac.ir
\end{minipage}

\vspace{6pt}
\begin{minipage}[t]{0.6\textwidth}
\centering
Hadi Moradi\\
Advanced Robotics and Intelligent Systems Laboratory\\
School of Electrical and Computer Engineering\\
University of Tehran\\
Tehran, Iran\\
moradih@ut.ac.ir
\end{minipage}
}

\maketitle

\begin{abstract}
When discussing indoor localization using deep learning models, one of the most crucial aspects to consider is interpretability alongside predictive accuracy, as these models often act as black boxes. While accurate localization plays a key role in indoor navigation, in store advertising, and search and rescue operations, an additional interpretability function not only allows us to understand what happens under the hood of these models but also provides meaningful information that could be used for other purposes. For example, in this paper, the information obtained from the added interpretability was used to prune the input feature set. In this study, we proposed a hierarchical deep learning model that predicts the location alongside physics informed decoders. Interestingly, these decoders learn an effective geometry of the access points from RSSI measurements and the user positions they are labelled with, without being provided with the access points' real positions during training. This learned geometry was further used to rank and prune the access points. On the UJIIndoorLoc dataset, this chained hierarchical model achieves a mean 3D error of 7.07\,m, which is 26\% to 36\% lower than baseline models, while previously published results on the same official split report 10.6\% to 31.0\% higher error. Furthermore, with 35\% and 50\% of the access points pruned, the model retains almost the same localization accuracy as the full model. The learned geometry also gives the AP coordinates that classical optimal design methods need but fingerprint databases do not record, so access points can be ranked by Fisher information without any surveyed AP position. On two further datasets, Tampere/TUT and UTSIndoorLoc, this geometry guided selection matches selectors built directly from the labelled data.
\end{abstract}


\section{Introduction}
\label{sec:intro}

GPS stops working the moment one walks through a doorway, and indoor localization is what fills that gap: a warehouse robot picking the fastest route between shelves and a hospital trying to locate a crash cart both need to know where they are indoors, often within seconds. Because most buildings already have WiFi infrastructure and no extra hardware is needed, WiFi signal has become a popular, inexpensive way to solve this problem. The most common method is fingerprinting, whereby an offline phase collects RSSI vectors at known positions to build a fingerprint database, and an online phase subsequently matches a new RSSI vector against it.

Deep learning brings a second concern alongside accuracy, which is interpretability. Deep fingerprinting models are treated as black boxes, with a RSSI vector fed in, a coordinate returned, and nothing in between open to inspection. This matters for trust, but it matters more for the aspect we pursue here, because if what the model learns internally is physically meaningful, that internal knowledge can be extracted and reused for other purposes entirely. Here it is used to decide which access points are worth keeping, so interpretability earns its place by supporting a concrete engineering decision rather than by producing a picture.

A further challenge is the number of access points (APs) involved. Modern buildings can have hundreds, making the fingerprint database large, the survey slow, and deep models heavy at inference time, so many studies select a small subset of APs that retains most of the accuracy. Classic selection methods rely on statistical scores such as detection rate, mutual information with position, or signal variance, but they carry no knowledge of where the AP physically is, since they consider only the RSSI numbers themselves.

AP selection need not remain a black box, because physics informed models embed a propagation model inside a neural network, so the network predicts position while simultaneously learning radio channel parameters such as AP location and path loss exponent. This AP location is initialized from the training position centroid and then learned from RSSI patterns and the training positions, never from the router's true coordinate, and we therefore call it effective geometry, since it represents where the network believes the signal originates, based on how the signal changes with position.

Our central idea is that a meaningful effective geometry should carry useful spatial information. We exploit it by pruning which APs we keep in the input, lowering survey and computation cost.

Our evidence is organized around three properties of the recovered geometry. It is faithful, in that the geometry behaves like real radio geometry rather than an arbitrary embedding that merely reconstructs RSSI, with a randomized control (same model, random frozen AP coordinates) as a check against exactly that. It is nearly free, in that the interpretable configuration costs no considerable amount of accuracy (7.07 m against 7.21--7.29 m for the ablations). It is useful, in that the recovered geometry can stand in for true AP coordinates inside classical geometry driven machinery, so a selection score computed from the geometry alone matches scores built directly from the labelled data, while leaving behind an artifact that no purely statistical score produces.

What remains is to test whether the learned geometry carries real information about the input features themselves, which access points matter and which do not. We measure it with metrics that were built for known sensor layouts, which let us score each WAP by relating the input features to the AP geometry the network has learned. Fisher information is one such metric (Section~\ref{sec:related}). The D optimal design criterion it implements is a classic tool from optimal experiment design, normally unavailable in WiFi fingerprinting because it requires knowing where the sensors are, and real AP coordinates are almost never published with a fingerprint database. The effective geometry supplies exactly that missing input. We therefore build the metrics over the learned geometry, use them to rank APs from most to least informative, and compare against 4 other pruning methods (random, detection count, mutual information, and a geometry + physics + detection rank ensemble we call multi signal consensus pruning, MSCP) at 4 pruning levels (20\%, 35\%, 50\%, 65\% of input features pruned), across 5 random seeds and on 3 independently collected datasets. A score read off the geometry alone that keeps pace with scores built from labels is evidence about the geometry, since it carries no information the geometry did not give it.

Contributions of this paper:
\begin{itemize}
\item The resulting model outperforms the classic baselines and every published result we could confirm on the official UJIIndoorLoc split (Section~\ref{sec:sota}).
\item We propose a physics informed encoder decoder network that learns building, floor, and 2D position through a differentiable path loss decoder, and that recovers an effective AP geometry with no real AP position label ever given.
\item The same decoder recovers a per building and per AP path loss exponent $n_{\mathrm{eff}}$ and a floor attenuation coefficient $\beta$ with no construction label given, a second layer of physical interpretability (Section~\ref{sec:physics-model}).
\item We provide a geometry diagnostic suite and a test time robustness analysis of pruned models under further AP removal and RSSI noise, applied across multiple independently collected datasets.
\end{itemize}

Section~\ref{sec:method} presents the model, its training, and how the recovered geometry is put to a D optimal design test; Section~\ref{sec:setup} the experimental setup; Section~\ref{sec:results} the results; and Section~\ref{sec:discussion} what they mean and where the approach can fail.

\section{Related Work}
\label{sec:related}

\textbf{Classic fingerprinting.} Radio map fingerprinting dates to RADAR~\cite{bahl2000radar}, which performs deterministic $k$ nearest neighbor matching against a calibrated database, and to Horus~\cite{youssef2005horus}, which instead applies probabilistic Bayesian inference over per location RSSI distributions. Consequently, this lineage underlies the WKNN/KNN baselines used in Section~\ref{sec:setup}.

\textbf{Deep learning for fingerprinting.} CNNLoc~\cite{song2019cnnloc} combines a stacked autoencoder with a CNN classifier. On UJIIndoorLoc specifically, more recent work pushes architecture complexity further through sequential deep learning (ISDL~\cite{mao2025isdl}), heterogeneous graph neural networks (MG-HGNN~\cite{wang2025mghgnn}), hierarchical stage wise training (Li et al.~\cite{li2025hierarchical}), attention based denoising autoencoders (Liu et al.~\cite{liu2025svaewganade}), area based localization (DeepLocBox~\cite{laska2021deeplocbox}), and federated hierarchical training (Etiabi et al.~\cite{etiabi2023federated}), and these are precisely the methods we compare against directly in Section~\ref{sec:sota}. Our encoder is likewise trained denoising style (Section~\ref{sec:physics-model}), but the reconstruction it serves is a physics decoder rather than a classifier, so the denoising objective is what keeps the latent informative enough for that decoder to fit a propagation model against. A separate GNN based line of work jointly learns AP selection and localization~\cite{wang2024gnnapselection}, and a related direction uses fine grained channel state information (CSI) rather than RSSI~\cite{liu2025gtcncsi}. However, all of these improve accuracy through architectural expressiveness alone.

\textbf{AP/feature selection.} Classic AP selection ranks or filters access points by statistical criteria computed directly on RSSI, such as maximum signal strength, entropy or information gain, and Kullback--Leibler divergence~\cite{laitinen2016apselection}, and this is the same family of methods (detection count, mutual information) that we compare against in Section~\ref{sec:setup}. A related, more general line of work frames sensor subset selection itself as a convex relaxation of the Fisher information objective~\cite{joshi2009sensor,kiefer1959optimum,pukelsheim2006optimal}, which is the classical perspective the criterion we apply to the learned geometry (Section~\ref{sec:fisher-selection}) comes from.

\textbf{Interpretability in deep models.} Interpretability methods are conventionally divided into two families. Post hoc methods leave the model untouched and attribute its predictions afterwards, as in LIME~\cite{ribeiro2016lime} and SHAP~\cite{lundberg2017shap}, so the explanation is produced by a second procedure, and it is not guaranteed to describe what the network actually computed. Intrinsic methods instead constrain the model so that some part of it is meaningful by construction, an approach argued for by Rudin~\cite{rudin2019stop} on the grounds that an explanation which can be wrong about its own model is a poor basis for a consequential decision. Physics informed learning~\cite{karniadakis2021pinn} belongs to the second family, since the embedded physical law forces a subset of the parameters to carry a defined physical meaning. Our work is positioned there, since the interpretable artifact is not attributed after training but is a parameter set the network is obliged to fit, and it is evaluated not by inspection but by whether a downstream decision can be made from it. To our knowledge, interpretability has not previously been treated as the design objective in WiFi fingerprinting, where the deep learning literature cited above is organized almost entirely around accuracy.

\textbf{Physics informed localization.} A separate line of work embeds an explicit propagation model, typically the log distance path loss law~\cite{rappaport2002wireless}, inside the localization pipeline. In that work, however, AP coordinates and channel parameters are usually provided in advance or fit as a separate calibration step, and are not learned jointly and end to end from RSSI and user positions with no AP coordinate ever given. The closest prior attempts to recover AP location without ground truth instead use classical multidimensional scaling on RSSI dissimilarity~\cite{koo2012serendipity} or substitute inertial navigation trajectories for a manual survey~\cite{zhuang2015autonomous}. Neither learns AP geometry jointly with a differentiable physics decoder inside an end to end localization network, as we do here.

None of these threads offers what this paper does: an end to end physics informed model that recovers an effective, physically sensible AP geometry from RSSI and user positions, then exposes that geometry itself, not merely the accuracy it produces, as an interpretable artifact from which a downstream selection decision is made.

\section{Method}
\label{sec:method}

\subsection{Problem Setting}

Each sample is a pair $(x, m)$, with $x \in \mathbb{R}^{A}$ the RSSI vector over $A$ access points and $m \in \{0,1\}^{A}$ the detection mask, where $m_j = 1$ when AP $j$ was heard and $m_j = 0$ when it was not. The label is $y = (b, f, p)$, with building $b \in \{1,\dots,B\}$, floor $f \in \{1,\dots,F\}$ and planar position $p = (p_x, p_y) \in \mathbb{R}^{2}$, and the task is to learn the map $(x, m) \mapsto y$.

We evaluate on three independently collected fingerprint databases, all encoding a non detection with the same sentinel value so that a single loader serves all three. UJIIndoorLoc~\cite{ujiindoorloc2014} is the primary one, with $A = 520$, $B = 3$, $F = 5$, and an official split of 19{,}937 training and 1{,}111 test samples collected from many phones over a long time period. Tampere/TUT ($A = 992$, $B = 1$, $F = 5$) and UTSIndoorLoc ($A = 589$, $B = 1$, $F = 16$) are used in Section~\ref{sec:wap-selection-results} to test whether the selection behavior transfers to other sites.

\subsection{Physics Informed Model}
\label{sec:physics-model}

We propose \emph{PhysicsChainedNet}, an encoder decoder network with three components, described in turn below: an encoder, a chain of hierarchical prediction heads, and two RSSI decoders.

\begin{figure*}[t]
\centering
\includegraphics[width=\textwidth]{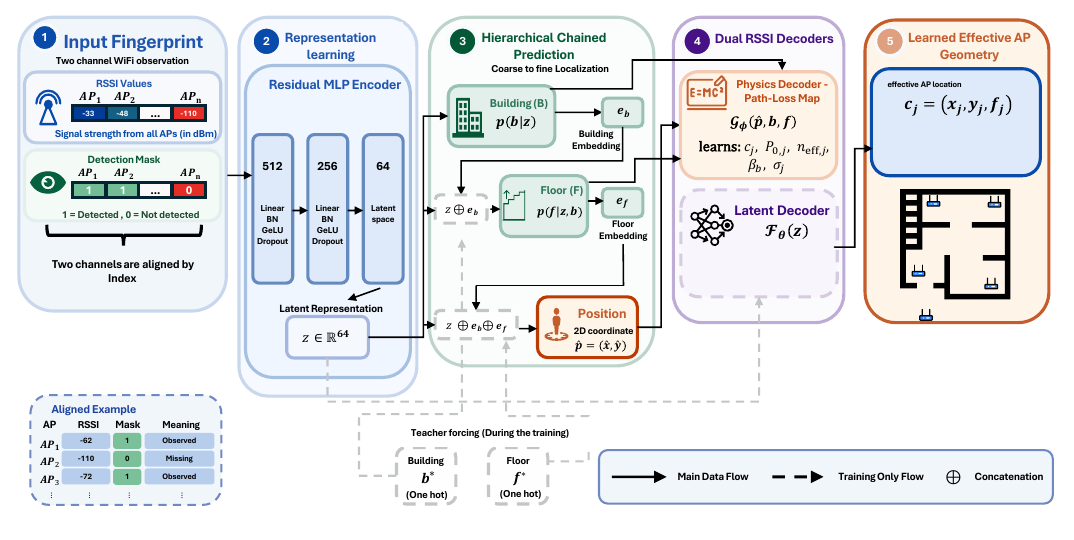}
\caption{\emph{PhysicsChainedNet}. The two channel input (normalized RSSI and detection mask, aligned by index) passes through the residual MLP encoder to the latent vector $z$, then through the chained building $\to$ floor $\to$ position heads. Two decoders reconstruct RSSI from different representations: the physics decoder from the predicted position $\hat p$ and the building/floor distributions, and the latent decoder from $z$. Only the physics decoder carries interpretable radio parameters, and its learned AP coordinates $c_j$ are the effective geometry of panel~5. Training time paths beyond teacher forcing are omitted; see Section~\ref{sec:training-procedure}.}
\label{fig:architecture}
\end{figure*}

\begin{figure}[t]
\centering
\includegraphics[width=\columnwidth]{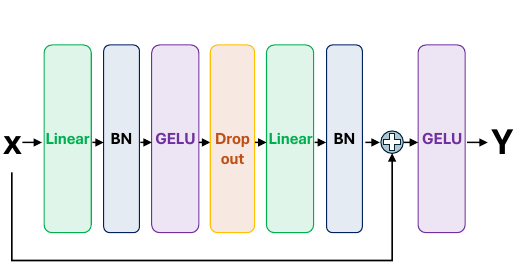}
\caption{The residual block used throughout the encoder stack. The skip path carries the input around the two linear layers and rejoins before the final activation.}
\label{fig:resblock}
\end{figure}

\textbf{Encoder.} The encoder maps the two channel input (normalized RSSI and detection mask) to a latent vector through a residual MLP stack (input $\to$ 512 $\to$ 256 $\to$ 64) built from the block of Fig.~\ref{fig:resblock}.

\textbf{Hierarchical chained head.} Three heads predict building, floor, and position in sequence rather than independently (Fig.~\ref{fig:architecture}, panel~3), since knowing the building helps predict the floor, and knowing both helps predict position. Each head's prediction is turned into a 32 dimensional embedding that conditions the next head in the chain.

Which embedding is used depends on \emph{teacher forcing}~\cite{williams1989teacherforcing}: during training, the model takes the true label's embedding with probability equal to the current teacher forcing ratio. Otherwise it takes a soft blend, using the building head's softmax output as weights over the 3 building embeddings. The blend is preferred to a hard argmax because it is differentiable, so the downstream heads still receive a gradient, and it carries uncertainty forward rather than committing to a single guess. At test time the ratio is zero, so the blend is always used. Section~\ref{sec:training-procedure} gives the schedule.

\textbf{Physics decoder.} The key component is the physics decoder, which we refer to as the path loss map (Fig.~\ref{fig:architecture}, panel~4). For predicted position $\hat p$, building $b$ and floor $f$, it computes expected RSSI for AP $j$ as
\begin{equation}
\hat\mu_j(\hat p) = P_0 - 10\, n_{\mathrm{eff},j} \log_{10}\!\left(\frac{d_j}{d_0}\right) - \beta\,|\Delta f_j|,
\end{equation}
where $d_j = \max(\|\hat p - c_j\|, d_0)$ is the distance between predicted position $\hat p$ and \emph{learned} AP coordinate $c_j$, $P_0$ is reference power (the expected RSSI right at the AP at distance $d_0=1$\,m), $\beta$ is a floor attenuation coefficient, the extra signal loss per floor of vertical separation $|\Delta f_j|$, and $n_{\mathrm{eff},j}$ is the \emph{path loss exponent} for AP $j$.

$n_{\mathrm{eff}}$ sets how fast power falls with $\log_{10}(\mathrm{distance})$, about 2 in free space and 2 to 5 indoors, and is learned per building and adjusted per AP (Section~\ref{sec:hyperparams}) because obstruction differs between both.

These, together with a per AP shadowing variance $\sigma_j^2$, are all learnable and are trained with the encoder using a reconstruction loss. Each is bounded to a physically sensible range by a smooth sigmoid style mapping, namely $[-110, 0]$\,dBm for $P_0$, $[1.1, 5.0]$ for $n_{\mathrm{eff},j}$, $[0, 30]$\,dB per floor for $\beta$ and $[2, 15]$\,dBm for $\sigma_j$, so the physics layer cannot drift into a nonsense regime like negative attenuation.

\textbf{Latent decoder.} The second decoder reconstructs RSSI from the latent vector $z$ rather than from the predicted position, and carries no physical parameters at all. The encoder input is deliberately corrupted during training: a mask aware augmentation drops each access point with probability $0.15$ and adds Gaussian noise of scale $0.02$ to the channels that survive, while both reconstruction losses are scored against the \emph{uncorrupted} RSSI. The encoder and this decoder therefore form a denoising autoencoder, with a corruption model that mirrors the way access points genuinely drop out between survey and query. Section~\ref{sec:loss} explains why both reconstruction terms are kept.

\begin{figure}[t]
\centering
\includegraphics[width=0.85\linewidth]{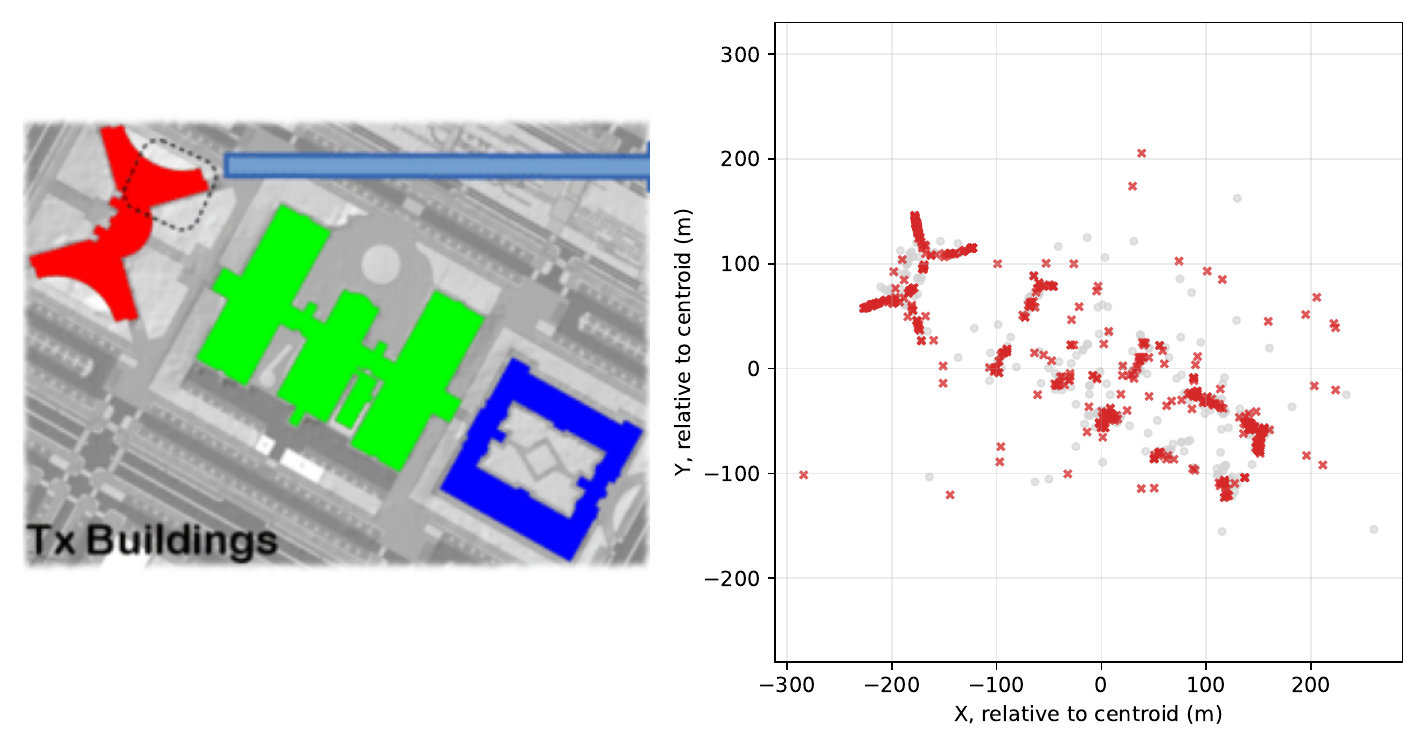}
\caption{Left: campus map, the transmitter (Tx) buildings colored (map from \cite{ujiindoorloc2014}). Right: learned effective AP coordinate (red $\times$) versus the centroid based initial guess (grey dot), from a model given no real building shape or AP location (quantitative faithfulness check in Section~\ref{sec:geom-diag}).}
\label{fig:ap-map}
\end{figure}

\subsection{Training Procedure}
\label{sec:training}
\label{sec:training-procedure}

\begin{figure*}[t]
\centering
\includegraphics[width=\textwidth]{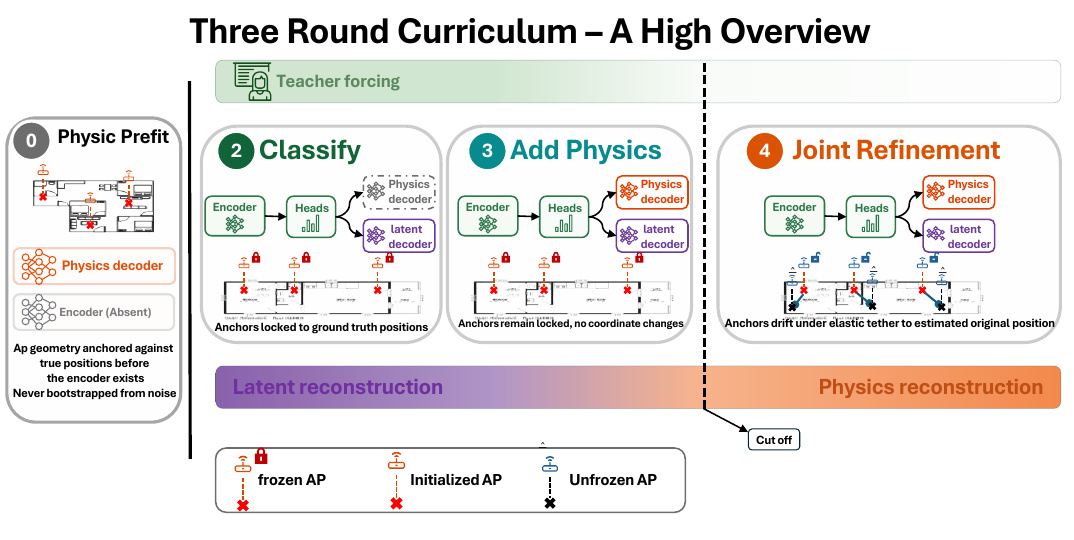}
\caption{The training procedure. Left of the vertical rule, the physics prefit fits the decoder's own parameters against true training positions, before the encoder is trained at all. The three frames to its right are the curriculum rounds, each drawing the same model schematic: a dashed grey decoder outline marks an objective that is not yet engaged, a solid outline one that is active. Access point anchor state follows the legend, and in the final round the anchors become trainable and are drawn drifting from their prefit positions. The vertical rule marks the boundary between the second and third rounds, where the teacher forcing ratio reaches zero. Table~\ref{tab:curriculum} gives the corresponding weights.}
\label{fig:curriculum}
\end{figure*}

\begin{table}[t]
\centering
\caption{Settings that change from round to round. Loss weights are those of the tuned configuration (Section~\ref{sec:hpsearch}); a dash marks a term inactive in that round. The physics prefit preceding round~1 runs 17 epochs at learning rate 0.0313.}
\label{tab:curriculum}
\small
\begin{tabular}{lccc}
\toprule
 & Round 1 & Round 2 & Round 3 \\
\midrule
Share of epoch budget & 0--25\% & 25--50\% & 50--100\% \\
\addlinespace[2pt]
Position $w_{\mathrm{pos}}$        & 5    & 8    & 25 \\
Floor $w_{\mathrm{floor}}$         & 20   & 15   & 20 \\
Building $w_{\mathrm{build}}$      & 15   & 10   & 8 \\
Physics recon.\ $w_{\mathrm{phy}}$ & 0.83 & 8.27 & 9.93 \\
Latent recon.\ $w_{\mathrm{rec}}$  & 1.0  & 0.8  & 0.5 \\
\addlinespace[2pt]
AP coordinates                     & frozen & frozen & trainable \\
AP drift $w_{\mathrm{drift}}$      & ---  & ---  & 0.211 \\
AP boundary $w_{\mathrm{bnd}}$     & ---  & ---  & 0.1 \\
\addlinespace[2pt]
Teacher forcing ratio & $1.0\!\to\!0.5$ & $0.5\!\to\!0.0$ & $0$ \\
\bottomrule
\end{tabular}
\end{table}

We structure training in three stages: a one off physics warm up before any curriculum round begins, the three round curriculum~\cite{bengio2009curriculum} itself, and the teacher forcing schedule used by the chained heads (Section~\ref{sec:physics-model}). This procedure applies identically whether training the initial full WAP model or retraining on a pruned WAP subset (Section~\ref{sec:fisher-selection}). Figure~\ref{fig:curriculum} shows the procedure end to end, and Table~\ref{tab:curriculum} gives the settings that change from round to round.

\textbf{Physics prefit.} Before any curriculum round begins, we fit the physics decoder's own parameters, including the AP coordinates $c_j$, directly against the true training positions $p$, using the same physics reconstruction likelihood given in Section~\ref{sec:loss}, but evaluated at the ground truth position instead of one the encoder has not yet learned to predict. The physics decoder is therefore never bootstrapped jointly with the encoder from an arbitrary starting point: by the time curriculum round 1 begins it already represents a physically grounded, if coarse, radio geometry anchored to real coordinates. This matters because an AP whose anchor starts in the wrong building is hard to move across the site later, and the drift penalty of Section~\ref{sec:loss} measures displacement from exactly this starting point. We repeat this prefit on the retained WAP set before every retraining that follows WAP pruning.

\textbf{Curriculum rounds.} The three rounds differ in which objectives are active and in whether the AP coordinates may move, as set out in Figure~\ref{fig:curriculum} and Table~\ref{tab:curriculum}; the coordinates are released only in the final round, and only under the AP drift penalty (Section~\ref{sec:loss}). This order lets the network learn a coarse position sense before coadapting with the already anchored radio geometry, rather than both sides starting from scratch together.

\textbf{Teacher forcing schedule.} Left to itself from epoch 0, the chain would build floor and position predictions on nearly random building/floor guesses. Following a scheduled sampling decay~\cite{bengio2015scheduled}, the teacher forcing ratio decays linearly from 1.0 (always the true label) to 0.0 (Table~\ref{tab:curriculum}), so the model is only gradually forced to rely on its own predictions, as it must at test time where no true label is available. The decay reaches zero exactly at the boundary between rounds 2 and 3, so the model stops being shown true labels at the same point its AP coordinates become trainable.

\subsection{Training Objective}
\label{sec:loss}

We define the total training loss as a weighted sum of six terms:
\begin{equation}
\mathcal{L} = \mathcal{L}_{\mathrm{loc}} + w_{\mathrm{rec}}\mathcal{L}_{\mathrm{rec}}^{\mathrm{nn}} + w_{\mathrm{phy}}\mathcal{L}_{\mathrm{rec}}^{\mathrm{phy}} + \mathcal{L}_{\mathrm{drift}} + \mathcal{L}_{\mathrm{floor}} + w_{\mathrm{bnd}}\mathcal{L}_{\mathrm{bnd}}.
\end{equation}

\textbf{Localization loss} $\mathcal{L}_{\mathrm{loc}}$ combines position, floor and building error:
\begin{align}
\mathcal{L}_{\mathrm{loc}} = {}& w_{\mathrm{pos}}\big(\mathcal{L}_{\mathrm{pos}} + w_{3D}\mathcal{L}_{3D}\big) \nonumber \\
& + w_{\mathrm{floor}}\,\mathrm{CE}(l_f, f) + w_{\mathrm{build}}\,\mathrm{CE}(l_b, b),
\end{align}
where $\mathcal{L}_{\mathrm{pos}}$ is Smooth L1 (Huber) loss~\cite{huber1964robust} between predicted and true 2D position, and $\mathrm{CE}$ is the standard classification cross entropy loss~\cite{goodfellow2016deep}. $\mathcal{L}_{3D}$ is an auxiliary "expected 3D" term: it turns the floor logit into a probability over floors, then computes the expected squared vertical error under this whole distribution (not just the single most likely floor), so a model that is unsure between two floors still gets penalized correctly:
\begin{equation}
\mathcal{L}_{3D} = \frac{1}{H}\sqrt{\|\hat p - p\|^2 + H^2\sum_k \pi_k (k - f)^2},
\end{equation}
with $\pi_k=\mathrm{softmax}(l_f)_k$ the predicted probability of candidate floor $k$, $f$ the true floor, and $H$ the floor to meter height constant (4\,m).

Smooth L1 is quadratic below a threshold $\delta$ and linear above it, so a few large errors cannot dominate the batch gradient. The expected floor term also gives the position head a useful gradient while the floor head is still unsure, which is common early in training.

\textbf{Reconstruction loss} $\mathcal{L}_{\mathrm{rec}}^{\mathrm{nn}}$ and $\mathcal{L}_{\mathrm{rec}}^{\mathrm{phy}}$ both compare reconstructed RSSI against the true input, separately for detected APs, at weight 1, and non detected APs, at weight $w_{\mathrm{empty}}=0.1$ or a censored likelihood explained next. Both terms exist together, not just the physics one, for two reasons. The first is optimization: the physics decoder is tightly constrained from the start, so the latent decoder supplies an easier reconstruction signal before the AP geometry becomes meaningful, and without this second path training is more likely to converge to a poor, arbitrary geometry. The second is representational, and it applies throughout training rather than only early: the physics path reaches the input only through the two dimensional $\hat p$ and the building and floor distributions, so on its own it places no pressure on $z$ to retain anything beyond what the three heads need. Reconstructing all $A$ channels from $z$ keeps the latent sufficient for the RSSI pattern itself, which is why the weight on this term decays but never reaches zero (Section~\ref{sec:hyperparams}). For the physics path we evaluate $\mathcal{L}_{\mathrm{rec}}^{\mathrm{phy}}$ as a Gaussian negative log likelihood in dBm space for detected APs,
\begin{equation}
\ell_j^{\mathrm{det}} = \frac{(\mathrm{RSSI}_j - \hat\mu_j)^2}{2\sigma_j^2} + \log\sigma_j,
\end{equation}
using the learned per AP shadowing $\sigma_j$ of the physics decoder, plus a \emph{censored} likelihood~\cite{tobin1958estimation} for non detected APs. An AP that was not seen just means its true signal is somewhere below the detection threshold $\tau=-110$\,dBm, so we penalize the model only if it predicts a signal clearly \emph{above} threshold, using a one sided relu squared penalty, not a hard target value:
\begin{equation}
\ell_j^{\mathrm{cens}} = \mathrm{ReLU}^2\big(\hat\mu_j - \tau\big),
\end{equation}
The learned per AP $\sigma_j$ lets the Gaussian likelihood down weight APs already known to be noisy, instead of one fixed weight over penalizing them all equally.

\textbf{Soft geometry constraints.} Once AP coordinates become trainable in round 3, nothing in the reconstruction loss alone stops them from drifting to a degenerate or out of range solution that nonetheless still reconstructs RSSI well, which is an identifiability problem common to any learned embedding. Three penalties active only in round 3 keep the geometry physically plausible without freezing it completely, all sharing the same one sided squared penalty mechanic: $\mathrm{ReLU}^2(\cdot)$ is zero while a quantity stays inside its valid range and grows with the square of the violation once it leaves. \emph{AP drift} $\mathcal{L}_{\mathrm{drift}}$ penalizes the average distance from each AP's initial (prefit) coordinate,
\begin{equation}
\mathcal{L}_{\mathrm{drift}} = w_{\mathrm{drift}} \cdot \frac{1}{|\mathcal{J}|}\sum_{j\in\mathcal{J}} \|c_j - c_j^{(0)}\|,
\end{equation}
so positions can adapt but not wander arbitrarily far. The remaining two apply that mechanic directly: \emph{floor coordinate} $\mathcal{L}_{\mathrm{floor}}$ holds the AP's learned floor value $\zeta_j$ inside $[0,1]$ (0 = ground floor, 1 = top floor), and \emph{boundary} $\mathcal{L}_{\mathrm{bnd}}$ holds $(x,y)$ inside the training footprint plus a 50\,m margin, so APs cannot drift to some far away point just to lower reconstruction error slightly. Together these three terms are what later lets us read the coordinates as an interpretable map (Fig.~\ref{fig:ap-map}).

\subsection{Hyperparameter Search}
\label{sec:hpsearch}
\label{sec:hyperparams}

None of the weights and bounds introduced above were chosen by hand. They come instead from the search described in this section. The values that matter for interpreting the model, in particular the physical bounds of Section~\ref{sec:physics-model} and the curriculum boundaries of Section~\ref{sec:training}, are quoted in the text where they matter, and the remaining optimization settings are reported in the supplementary material.

We tune hyperparameters with a multi fidelity Gaussian Process search (a Gaussian Process Regressor with a Matern kernel plus a white noise term~\cite{rasmussen2006gpml}), treating the number of training epochs as an extra input dimension so that cheap low epoch trials screen configurations before full length runs. Configurations are promoted through 100, 300, 600 and 1000 epochs over 48 training runs in total, and the best is confirmed across all 5 seeds. Only the optimization hyperparameters are tuned (learning rate, weight decay, dropout, noise scale, physics loss weight, AP drift weight, expected 3D weight, prefit learning rate and epochs, and Smooth L1 $\delta$), not the architecture or the choice of ablation. The search scores on a held out slice of \emph{only} the training data, never touching the official validation split.

\subsection{Probing the Learned Geometry with D Optimal Design}
\label{sec:fisher-selection}

The recovered geometry is only meaningful if it supports the geometric reasoning that true access point coordinates support. We therefore take a standard criterion from optimal experiment design, apply it to the learned coordinates unchanged, and let the quality of the resulting selection report back on the geometry. The criterion quantifies how much position information AP $j$ carries at a given user position $p$, which is what Fisher information measures. Every quantity in what follows is read off the learned physics decoder, so the criterion below is computable on any fingerprint database, whether or not the true AP coordinates were ever recorded. The Jacobian is
\begin{equation}
J_j(p) = \frac{\partial \hat\mu_j(p)}{\partial p} \in \mathbb{R}^2
\end{equation}
obtained by automatic differentiation of the learned path loss function. For AP subset $S$, the position Fisher information matrix at $p$ is
\begin{equation}
F_S(p) = \sum_{j \in S} \frac{q_j(\text{detect} \mid p)}{\sigma_j^2}\, J_j(p)\, J_j(p)^\top,
\end{equation}
where $q_j(\text{detect}\mid p)$ is detection probability from the learned detection threshold model. A bigger $\log\det F_S(p)$ means the AP subset $S$ constrains the position better at $p$ (this is the classic D optimal design criterion~\cite{kiefer1959optimum,pukelsheim2006optimal}, but here it uses \emph{learned}, not known, AP geometry).

The selection objective is the standard one, the average log determinant of the Fisher matrix over $N$ sample positions:
\begin{equation}
Q(S) = \frac{1}{N} \sum_{p} \log\det\big(F_S(p) + \epsilon I\big),
\end{equation}
where $\epsilon$ is a small constant added for numerical stability.

APs are ordered by greedy forward selection, the standard construction for this objective~\cite{wynn1970sequential}: starting from the empty set, each step adds the AP giving the biggest gain in $Q$, batched over positions for tractability. The ranking is therefore \emph{nested}, so the set retained at any prune level is a prefix of the set retained at a shallower one and no level needs recomputing.

Once a WAP subset is selected this way, whether by Fisher information ranking or by any of the compared baseline pruning methods, we do not simply mask the unused APs at inference on the already trained model. Instead we retrain \emph{PhysicsChainedNet} from scratch on only the retained WAPs, redoing the physics prefit step before this retraining, using the same training procedure and curriculum described in Section~\ref{sec:training-procedure}.

\section{Experimental Setup}
\label{sec:setup}

With the model, training procedure and selection method now fully specified, the rest of this section fixes exactly how we test them: the dataset and split, the baselines we compare against, the pruning methods the geometry based criterion is measured against, and the metrics used to score all of it.

\subsection{Dataset and Protocol}
\label{sec:dataset-protocol}
We use UJIIndoorLoc~\cite{ujiindoorloc2014} official split: 19{,}937 samples for train and 1{,}111 samples for test, which is the harder protocol because train and test were collected at different times. No detection rate prefiltering is applied, so all 520 WAPs of the dataset are retained as the input feature set, and every kept WAP count and pruned percentage reported below is computed against that full set of 520.

\subsection{Baselines}
\label{sec:baselines}
We compare against classic fingerprinting baselines: Weighted K Nearest Neighbor (WKNN, $k$=7), K Nearest Neighbor ($k$=1) and a simple feed forward MLP, all trained and tested on the same split.

The MLP baseline uses default hyperparameters. We additionally compare against published results on the same official split (Section~\ref{sec:sota}).

\subsection{Pruning Methods Compared}
We compare five selection strategies at the same prune levels. Two read the learned geometry, namely the \emph{Fisher} criterion of Section~\ref{sec:fisher-selection} and \emph{MSCP}, which uses it as one of three inputs; the remaining three do not see it at all. \emph{Random} keeps a uniformly random AP subset, with 3 independent redraws per nonzero prune level. \emph{Detection count} keeps the APs seen most often across the training set. \emph{Mutual information} ranks APs by mutual information between RSSI and each of three targets (the two position coordinates, and the combined building$\times$floor class label) via a nearest neighbor estimator. \emph{MSCP} is a hybrid selector that rank normalizes and averages three per AP scores (learned geometry, physics reconstruction residual, inverted detection count) into one combined score. All methods use 4 prune levels (20\%, 35\%, 50\%, 65\% pruned) plus the full model as a common reference, over 5 random seeds.

\subsection{Metrics}
We report mean, median, P90 and P95 of 3D localization error (building height converted with a fixed floor height), building accuracy and floor accuracy. For method comparison we run paired significance tests (paired t test and the Wilcoxon signed rank test~\cite{wilcoxon1945individual}) across seeds, with Holm correction for multiple comparisons~\cite{holm1979simple} and report Cohen's $d_z$ effect size.

\section{Results}
\label{sec:results}

The results follow the order of proof set out in Section~\ref{sec:intro} (faithful, free, useful): Section~\ref{sec:baselines-results} establishes task competitiveness, Section~\ref{sec:ablations} the accuracy free cost, Section~\ref{sec:geom-diag} faithfulness, and Section~\ref{sec:wap-selection-results} onward the geometry's usefulness for selection, degradation, robustness, and transfer to two further datasets.

\subsection{Comparison with Baselines and Published Results}
\label{sec:baselines-results}

\begin{figure}[t]
\centering
\includegraphics[width=\linewidth]{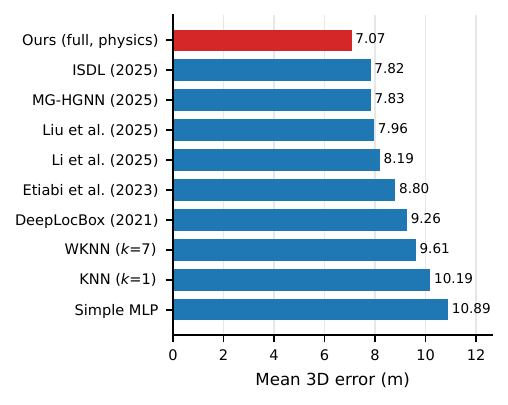}
\caption{Mean 3D error, our full model versus both the classic baselines (Table~\ref{tab:baselines}) and published results on the official UJIIndoorLoc split (Table~\ref{tab:sota}), ours highlighted in red, sorted ascending.}
\label{fig:baselines-sota}
\end{figure}

As shown in Table~\ref{tab:baselines} and Fig.~\ref{fig:baselines-sota}, our full physics informed model clearly outperforms the classic baselines. Mean 3D error drops from 9.61\,m (WKNN, $k$=7) to 7.07\,m, a 26.4\% improvement, and P90 error drops 17.9\%. Floor accuracy also improves, 93.4\% versus 88.6\% for the nearest neighbor method.

\begin{table}[t]
\centering
\caption{Full model vs. classic baselines. Our model: mean $\pm$ std over 5 seeds.}
\label{tab:baselines}
\small
\setlength{\tabcolsep}{3pt}
\begin{tabular*}{\columnwidth}{@{\extracolsep{\fill}}lccc@{}}
\toprule
Model & Mean 3D (m) & P90 (m) & Floor Acc. \\
\midrule
\textbf{Ours (full, physics)} & \textbf{7.07 $\pm$ 0.06} & \textbf{15.97} & \textbf{93.38\%} \\
WKNN ($k$=7)      & 9.61  & 19.45 & 88.57\% \\
KNN ($k$=1)       & 10.19 & 21.17 & 88.57\% \\
Simple MLP        & 10.89 & 21.58 & 91.54\% \\
\bottomrule
\end{tabular*}
\end{table}

\label{sec:sota}
Many UJIIndoorLoc papers reshuffle or resplit the training data (e.g. 70:30 or 90:10) instead of using the official split, since the true test labels were historically released only to IPIN/EvAAL participants, and this yields easier, optimistically biased numbers not comparable to an official split result. This is why we keep the harder official split throughout, and, as a sanity check, a shuffled 80/20 split of our own pooled data drops mean 3D error to 5.06\,m (versus 7.07\,m official), confirming the real distribution shift (time, devices, conditions) that a shuffled split erases.

\begin{table}[t]
\centering
\caption{Comparison with published results on the official UJIIndoorLoc split. The numbers are from the referenced papers.}
\label{tab:sota}
\scriptsize
\setlength{\tabcolsep}{3pt}
\begin{tabular*}{\columnwidth}{@{\extracolsep{\fill}}lcccc@{}}
\toprule
Method & Mean 3D (m) & Floor Acc. & Build. Acc. & vs. Ours \\
\midrule
\textbf{Ours (full, physics)}                & \textbf{7.07} & \textbf{93.38\%} & \textbf{99.93\%} & \textbf{-} \\
ISDL~\cite{mao2025isdl}                      & 7.82 & 95.05\% & 100\% & +10.6\% \\
MG-HGNN~\cite{wang2025mghgnn}                & 7.83 & 94.24\% & 100\% & +10.7\% \\
Liu et al.~\cite{liu2025svaewganade}         & 7.96 & 94.70\% & 100\% & +12.6\% \\
Li et al.~\cite{li2025hierarchical}          & 8.19 & 93.34\% & 100\% & +15.8\% \\
Etiabi et al.~\cite{etiabi2023federated}     & 8.80 & 94.87\% & 99.90\% & +24.5\% \\
DeepLocBox~\cite{laska2021deeplocbox}        & 9.26 & 92.12\% & 99.56\% & +31.0\% \\
Nowicki \& W.~\cite{nowicki2017loweffort}    & - & \multicolumn{2}{c}{92\% (combined)} & - \\
\bottomrule
\end{tabular*}
\end{table}

Our model reports the lowest mean 3D error among these confirmed official split comparators, but not on every metric. Floor accuracy (93.38\%) edges past DeepLocBox and Li et al., yet trails ISDL by nearly 2 points and MG-HGNN, Liu et al., and Etiabi et al. by roughly a point each.

\subsection{Physics Ablation Study}
\label{sec:ablations}

Before the recovered geometry is examined or used, the question of what it costs must be settled. We therefore investigate whether the physics component matters for accuracy by removing or degrading individual physics components and retraining from scratch. This study includes all 5 seeds for all 5 conditions, and Table~\ref{tab:ablation} reports the results.

\begin{table}[t]
\centering
\caption{Physics ablation, mean 3D error (m), mean $\pm$ std over 5 seeds.}
\label{tab:ablation}
\scriptsize
\setlength{\tabcolsep}{3pt}
\begin{tabular*}{\columnwidth}{@{\extracolsep{\fill}}lcc@{}}
\toprule
Condition & Mean 3D (m) & Floor Acc. \\
\midrule
\textbf{Full model}    & 7.07 $\pm$ 0.06 & 93.38\% \\
Fixed AP geometry       & 7.21 $\pm$ 0.10 & 92.31\% \\
No censored likelihood & 7.22 $\pm$ 0.10 & 91.90\% \\
No physics loss        & 7.27 $\pm$ 0.14 & 91.76\% \\
Randomized AP geometry  & 7.29 $\pm$ 0.16 & 92.53\% \\
\bottomrule
\end{tabular*}
\end{table}

\textbf{What each ablation condition changes.} \emph{No physics loss} removes the physics reconstruction and AP drift terms and freezes the decoder's parameters at initialization, so the decoder is structurally present but never updated. \emph{No censored likelihood} replaces the one sided relu squared penalty for non detected APs (Section~\ref{sec:loss}) with the plain weighted squared error used for detected APs, discarding the ``below threshold, not known exactly'' information. \emph{Fixed AP geometry} holds the AP coordinates at centroid initialization through all 3 rounds instead of unfreezing them in round 3. \emph{Randomized AP geometry} freezes them likewise but starts from uniformly random in footprint coordinates, testing whether the geometry needs to be sensible at all. All four are retrained from scratch with the same curriculum and hyperparameters, each differing from the full model in one respect only.

Overall, the four ablated variants land inside a narrow band running from 7.21 to 7.29\,m, while the full model sits lower at 7.07\,m. None of the differences in Table~\ref{tab:ablation} is large, which follows from the way the pipeline is arranged rather than from the physics loss being inert. What the physics loss does produce is the learned AP geometry and the per AP path loss parameters, and it is exactly this geometry on which the D optimal test of Section~\ref{sec:fisher-selection} operates.

This makes sense given which parameters each loss term touches. The localization loss never involves the AP coordinates and the geometry specific terms never involve the encoder, and the two sides meet only through the shared physics reconstruction loss, so freezing or randomizing the geometry leaves accuracy unchanged. What the physics decoder adds, then, is an inspectable geometry rather than raw accuracy, already used for WAP pruning (Section~\ref{sec:fisher-selection}).

\subsection{Geometry Diagnostics}
\label{sec:geom-diag}

Whether the recovered geometry is faithful, rather than an arbitrary embedding that happens to reconstruct RSSI well, is the question the rest of the paper depends on, and the diagnostics below test it directly. All numbers come from the full model, except coordinate stability, which compares AP coordinates across independently trained runs.

\textbf{RSSI distance consistency.} For each AP, we correlate its predicted distance to the user against its measured RSSI on the held out test set, since a physically sensible geometry should show signal weakening as predicted distance grows, that is, a negative correlation. Across the 232 APs with sufficient test detections to evaluate, the median correlation is indeed $-0.43$, and 94.8\% of APs show a negative correlation, consistent with real path loss rather than noise.

\textbf{Randomized geometry control.} As a further control, we recompute the same reconstruction likelihood and correlation with the predicted position replaced by a uniformly random in footprint position at evaluation time only, never during training, and the random location likelihood is worse (58.35 vs.\ 51.06 nats for the true predicted position) with its median correlation collapsing to $0.01$, confirming that the negative correlation above comes from the model actually locating the AP rather than from the likelihood itself. Using instead the \emph{randomized AP geometry} ablation model from Table~\ref{tab:ablation} (frozen, random in footprint AP coordinates in place of the learned ones) yields a median correlation of only $-0.004$, with just 51.7\% of APs negative, close to the coin flip rate expected from a genuinely uninformative geometry. The geometry the physics decoder learns therefore performs real physical work rather than serving as an arbitrary label for the reconstruction loss to lean on.

\textbf{Coordinate stability across seeds.} Comparing the learned AP coordinates pairwise across all 5 seeds' full models, the median AP displacement between any two seeds is 6.17\,m, with $x$/$y$ correlation of 0.89/0.89 across seed pairs. The geometry is thus reproducible in its broad shape but not pixel identical between runs, which is precisely the behavior we would expect from an identifiability constrained, though not fully identifiable, embedding (Section~\ref{sec:loss}).

\textbf{Parameter saturation.} We check what fraction of each learnable physics parameter sits at the edge of its permitted range (Section~\ref{sec:physics-model}), since a parameter pinned at its bound suggests the bound, not the data, is doing the work. $P_0$, the per building exponent, and floor attenuation are essentially never saturated (0 to 0.8\%), but the per AP shadowing $\sigma_j$ is saturated for 13.3\% of APs and the exponent offset $n_{\mathrm{eff},j}$ for 32.1\%, so roughly a third of APs would prefer an exponent more extreme than $[1.1, 5.0]$ allows. Both this parameter and the per AP $\sigma_j$ enter the Fisher matrix of Section~\ref{sec:fisher-selection}, so saturation also distorts the weight those APs receive in the selection criterion. This qualifies the interpretability claim of Section~\ref{sec:intro}. For that saturated third the exponent "read off" is set by the interval rather than the data and should not be interpreted as a physical property at all. Our $n_{\mathrm{eff}}$ claim is accordingly restricted to the unsaturated majority. We assume the saturated third are access points with few training detections or unusually noisy readings, so that their exponent is fit from too little evidence and is pushed to whichever bound best absorbs the residual, but we have not tested this and it needs further investigation. Checking that assumption against per access point detection counts and residual variance, alongside widening the bounds with a matching stability check and a comparison against each building's known construction, is left to future work.

The geometry is faithful in the aggregate but not in every particular, so the artifact is treated in what follows as a design time instrument, which is the use the next section puts it to.

\subsection{WAP Selection Across Three Datasets}
\label{sec:wap-selection-results}

\textbf{Setup.} We evaluate WAP selection under the same isolation test on three independently collected datasets. The geometry only score, the Fisher score over the learned geometry, sees the recovered access point coordinates and the physics layer's uncertainty and nothing else. Random pruning is redrawn three times per prune level and averaged. All three datasets retrain \emph{PhysicsChainedNet} following the identical procedure of Section~\ref{sec:training-procedure} and are compared at the same four prune levels, 20\%, 35\%, 50\%, and 65\% of input features removed.

\begin{figure}[t]
\centering
\includegraphics[width=\linewidth]{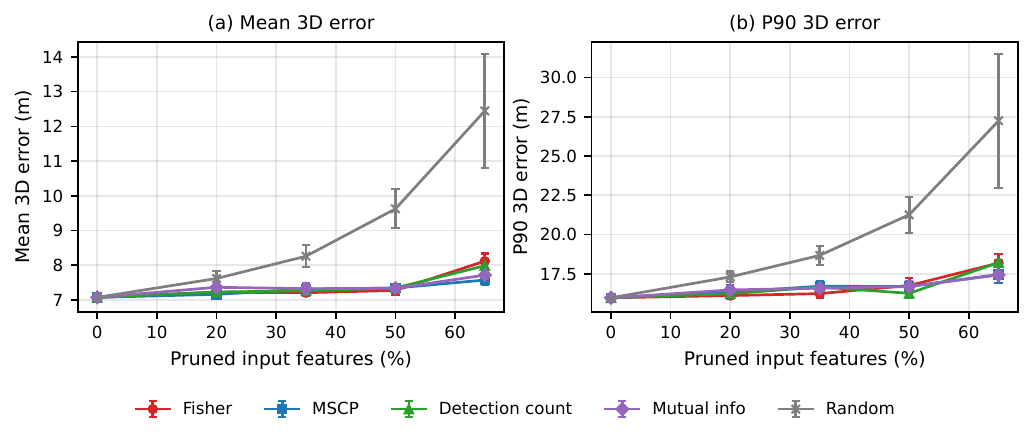}
\caption{(a) Mean 3D error and (b) P90 3D error versus pruned input features (\%), per selection method. The geometry only Fisher score tracks the selectors built from labels through moderate pruning.}
\label{fig:budget}
\end{figure}

\textbf{UJIIndoorLoc.} We report mean 3D error for every method at every prune level in Table~\ref{tab:wap} and Fig.~\ref{fig:budget}(a), and P90 error in Fig.~\ref{fig:budget}(b). Through the moderate pruning range the geometry only score is fully competitive with every alternative. It records the lowest error in the table at 35\% and 50\% pruned, alongside both purely statistical selectors and the hybrid, and beats random pruning by 5\% at 20\% pruned and by 35\% at 65\%. The margins over MSCP at these two levels are small, 1.4\% and 1.0\%, and do not survive Holm correction, so geometry only selection is the equal of the alternatives here. A score derived entirely from a geometry the model was never given matches scores built directly from the labelled data, while additionally leaving behind an inspectable access point position that none of the alternatives produce. MSCP is the strongest all round selector in the table, and the recovered geometry is one of the three signals it consumes.

\begin{table}[t]
\centering
\caption{UJIIndoorLoc: mean 3D error (m), mean $\pm$ std over 5 seeds, by method and pruned input features (\%). Fisher (in bold) is the geometry only selector, computed from the learned effective geometry and nothing else.}
\label{tab:wap}
\scriptsize
\setlength{\tabcolsep}{3pt}
\begin{tabular*}{\columnwidth}{@{\extracolsep{\fill}}lccccc@{}}
\toprule
Method & 0\% & 20\% & 35\% & 50\% & 65\% \\
\midrule
\textbf{Fisher} & \multirow{5}{*}{7.07$\pm$.06} & 7.22$\pm$.11 & \textbf{7.21$\pm$.10} & \textbf{7.27$\pm$.16} & 8.12$\pm$.25 \\
MSCP          &                                & \textbf{7.16$\pm$.07} & 7.31$\pm$.10 & 7.34$\pm$.04 & \textbf{7.57$\pm$.15} \\
Detect. count &                                & 7.33$\pm$.17 & 7.36$\pm$.06 & 7.44$\pm$.09 & 8.08$\pm$.12 \\
Mutual info   &                                & 7.36$\pm$.05 & 7.32$\pm$.18 & 7.34$\pm$.11 & 7.71$\pm$.15 \\
Random        &                                & 7.62$\pm$.15 & 8.26$\pm$.06 & 9.62$\pm$.28 & 12.45$\pm$1.16 \\
\bottomrule
\end{tabular*}
\end{table}

At the most aggressive prune level of 65\%, where only 182 of 520 access points are kept, this no longer holds. The geometry only score rises to 8.12\,m and is overtaken by MSCP at 7.57\,m, significant after Holm correction ($p$=0.032, Cohen's $d_z$=5.10), and also by mutual information at 7.71\,m and marginally by detection count at 8.08\,m. Because a selector using no geometry at all also overtakes it here, the reading is that geometry only selection has an operating range on 3D error and that 65\% lies outside it. The recovered geometry encodes where each access point sits and how informative it is about position, the right criterion while enough redundancy remains to spend. Once two thirds of the access points are gone, the binding constraint appears to be raw coverage, whether any well detected access point survives near a given region, and coverage is a statistic instead of a geometry. Separating the two would require a coverage constrained Fisher objective, which we leave to future work.

\label{sec:degradation}
Fig.~\ref{fig:budget}(a) shows the shape of this degradation: random pruning collapses, MSCP degrades most smoothly, and the geometry only score tracks the full model closely until 65\% pruned. A secondary finding is that floor accuracy actually \emph{improves} slightly under pruning for some methods. For instance, MSCP at 65\% pruned reaches 94.0\% floor accuracy versus 93.4\% for the full model, possibly because pruning acts as a light regularizer.

\textbf{Tampere/TUT.}
\label{sec:tampere}
We test whether the pruning behavior established on UJIIndoorLoc holds on a second, independently collected dataset, Tampere/TUT, crowdsourced from 21 Android devices with 992 access points across a single 5 floor building. A detection rate prefilter retains 779 of the 992 access points, and Table~\ref{tab:tampere} compares all five selection methods.

\begin{table}[t]
\centering
\caption{Tampere/TUT: (a) best 3D error (m) and (b) floor accuracy, by method and pruned input features (\%). Fisher (in bold) is the geometry only selector. Random averaged over three redraws.}
\label{tab:tampere}
\scriptsize
\setlength{\tabcolsep}{3pt}
\textbf{(a) Best 3D error (m)}\\[2pt]
\begin{tabular*}{\columnwidth}{@{\extracolsep{\fill}}lccccc@{}}
\toprule
Method & 0\% (full) & 20\% & 35\% & 50\% & 65\% \\
\midrule
\textbf{Fisher} & \multirow{5}{*}{7.36} & 7.30 & \textbf{7.36} & \textbf{7.33} & \textbf{7.51} \\
MSCP          &                         & 7.23 & 7.55 & 7.67 & 8.07 \\
Detect. count &                         & 7.30 & 7.47 & 7.75 & 8.15 \\
Mutual info   &                         & 7.25 & 7.40 & 7.50 & 8.17 \\
Random        &                         & 7.48 & 7.65 & 7.99 & 8.57 \\
\bottomrule
\end{tabular*}\\[6pt]
\textbf{(b) Floor accuracy}\\[2pt]
\begin{tabular*}{\columnwidth}{@{\extracolsep{\fill}}lccccc@{}}
\toprule
Method & 0\% (full) & 20\% & 35\% & 50\% & 65\% \\
\midrule
\textbf{Fisher} & \multirow{5}{*}{92.5\%} & 92.0\% & 92.0\% & 91.7\% & 91.6\% \\
MSCP          &                          & 92.1\% & 92.4\% & 91.8\% & 91.4\% \\
Detect. count &                          & 91.4\% & 92.7\% & 91.6\% & 90.9\% \\
Mutual info   &                          & 92.4\% & 92.2\% & 92.4\% & 91.7\% \\
Random        &                          & 91.5\% & 90.9\% & 90.2\% & 88.9\% \\
\bottomrule
\end{tabular*}
\end{table}

At 20\% pruned all four informed methods reach a 3D error at or below the full 779 access point model, 7.23 to 7.30\,m against 7.36\,m, while random pruning already trails it at 7.48\,m. Because random removes the same fraction without improving on the full model, the informed methods' gain is attributable to \emph{which} access points are removed, not to input size alone. From 35\% pruned onward the geometry only score leads every alternative, and its margin widens as pruning deepens. Its whole range spans 7.30 to 7.51\,m, so it never falls more than 0.15\,m below the full model even with two thirds of the access points removed, whereas every alternative gives up at least 0.84\,m over the same span. Floor accuracy separates the methods less, staying between 90.9\% and 92.7\% for all four informed selectors at every level, while random falls to 88.9\% at 65\% pruned.

\textbf{UTSIndoorLoc.}
\label{sec:uts}
We repeat the selection study on a third independently collected dataset, UTSIndoorLoc, gathered in a single 16 floor building with 589 access points and an official split of 9108 training and 388 test samples. A detection rate prefilter retains 556 of the 589 access points, and Table~\ref{tab:uts} compares all five selection methods. This site is the most demanding of the three for a geometry based argument, a single, tall, vertically dominated footprint in which a large share of the position information lies along the floor axis rather than across the plane, and in which the horizontal spread that the effective geometry is recovered from is correspondingly compressed.

\begin{table}[t]
\centering
\caption{UTSIndoorLoc: (a) best 3D error (m) and (b) floor accuracy, by method and pruned input features (\%). Fisher (in bold) is the geometry only selector. Random averaged over three redraws.}
\label{tab:uts}
\scriptsize
\setlength{\tabcolsep}{3pt}
\textbf{(a) Best 3D error (m)}\\[2pt]
\begin{tabular*}{\columnwidth}{@{\extracolsep{\fill}}lccccc@{}}
\toprule
Method & 0\% (full) & 20\% & 35\% & 50\% & 65\% \\
\midrule
\textbf{Fisher} & \multirow{5}{*}{7.30} & \textbf{7.42} & 7.83 & 8.28 & 8.42 \\
MSCP          &                        & 7.67 & 8.02 & 8.04 & 8.65 \\
Detect. count &                        & 7.45 & 7.79 & 7.73 & 8.34 \\
Mutual info   &                        & 7.51 & 7.38 & 7.51 & 7.94 \\
Random        &                        & 7.71 & 7.86 & 7.78 & 8.56 \\
\bottomrule
\end{tabular*}\\[6pt]
\textbf{(b) Floor accuracy}\\[2pt]
\begin{tabular*}{\columnwidth}{@{\extracolsep{\fill}}lccccc@{}}
\toprule
Method & 0\% (full) & 20\% & 35\% & 50\% & 65\% \\
\midrule
\textbf{Fisher} & \multirow{5}{*}{89.7\%} & 87.6\% & \textbf{88.4\%} & \textbf{88.1\%} & \textbf{85.6\%} \\
MSCP          &                          & 86.6\% & 87.6\% & 87.4\% & 82.5\% \\
Detect. count &                          & 88.9\% & 87.9\% & 88.1\% & 75.0\% \\
Mutual info   &                          & 89.2\% & 86.1\% & 85.3\% & 78.1\% \\
Random        &                          & 87.0\% & 88.7\% & 86.4\% & 79.5\% \\
\bottomrule
\end{tabular*}
\end{table}

The full 556 access point model reaches 7.30\,m with 89.7\% floor and 100\% building accuracy. At 20\% pruned the geometry only score gives the lowest 3D error of the five methods, 7.42\,m, within 1.7\% of the full model while a fifth of the input is discarded, and mutual information leads at the three deeper levels, and every informed method stays ahead of random pruning at 65\% pruned. The horizontal ordering among informed selectors is therefore budget dependent on this site.

Floor accuracy is where the recovered geometry declares itself, and on a 16 floor building it is the metric the site is built to test. The geometry only score holds 88.4\% and 88.1\% at 35\% and 50\% pruned, the best of any method at both levels, and at 65\% pruned it reaches 85.6\% while every alternative falls to between 75.0\% and 82.5\%. It is the only selector that stays within roughly four points of the full model once two thirds of the access points are gone, and it does so with a score that never sees a floor label. Vertical structure is a property of the geometry, and the selector that reads the geometry is the one that preserves it.

\textbf{What transfers.} Across three sites that differ in building count, floor count, footprint shape, and device population, the substitution of learned coordinates for surveyed ones holds. The sites disagree only at the deepest prune level, and informatively: coverage binds first where access points are scarcest, so on UJIIndoorLoc and UTSIndoorLoc the geometry only score is overtaken on 3D error at 65\% pruned, while on Tampere/TUT, which retains the largest access point set of the three, it leads from 35\% pruned onward. Where access points are plentiful the recovered geometry is the better criterion at every budget we tested.

\subsection{Robustness to AP Removal and RSSI Noise}
\label{sec:robustness}

Beyond the pruning time comparison earlier in this section, we additionally investigate what happens to an already trained, already pruned model at \emph{test time}: removing further APs it was never retrained without, and adding synthetic RSSI noise it never saw during training. We compare three already trained models: the full model, Fisher pruning at 35\% pruned, and mutual information pruning at 35\% pruned, the strongest purely statistical method at this prune level.

\begin{table}[t]
\centering
\caption{Test time robustness, mean 3D error (m): (a) additional AP removal at test time beyond the pruning already applied, and (b) synthetic RSSI noise (Gaussian, std in dBm), never seen during training.}
\label{tab:robustness}
\scriptsize
\setlength{\tabcolsep}{3pt}
\textbf{(a) Additional AP removal (\%)}\\[2pt]
\begin{tabular*}{\columnwidth}{@{\extracolsep{\fill}}lcccccc@{}}
\toprule
Condition & 0\% & 10\% & 20\% & 30\% & 40\% & 50\% \\
\midrule
Full (0\% pruned) & 7.11 & 7.44 & 8.12 & 8.46 & 9.27 & 10.44 \\
Fisher (35\% pruned) & 7.22 & 7.59 & 7.77 & 8.53 & 9.68 & 10.90 \\
Mutual info (35\% pruned) & 7.36 & 7.55 & 7.97 & 8.45 & 9.46 & 11.02 \\
\bottomrule
\end{tabular*}\\[6pt]
\textbf{(b) RSSI noise $\sigma$ (dBm)}\\[2pt]
\begin{tabular*}{\columnwidth}{@{\extracolsep{\fill}}lcccccc@{}}
\toprule
Condition & 0 & 0.01 & 0.02 & 0.05 & 0.10 & 0.15 \\
\midrule
Full (0\% pruned) & 7.11 & 7.11 & 7.14 & 7.56 & 8.76 & 10.16 \\
Fisher (35\% pruned) & 7.22 & 7.24 & 7.37 & 7.58 & 8.87 & 9.89 \\
Mutual info (35\% pruned) & 7.36 & 7.40 & 7.49 & 7.66 & 8.82 & 10.61 \\
\bottomrule
\end{tabular*}
\end{table}

All three conditions degrade smoothly under both stressors, and none shows a sudden collapse or cliff, which is itself a useful negative result, since an already deployed pruned model does not become brittle simply because it was pruned. Under further AP removal (Table~\ref{tab:robustness}a), the two 35\% pruned models degrade somewhat faster in relative terms than the full model, unsurprisingly, given that they have less redundancy left to lose. The comparison between them, however, is the more interesting part. Mutual information pruning degrades \emph{less} than Fisher pruning at every removal level except 20\%, ultimately ending up ahead again by 50\% removal (49.7\% versus 50.9\% relative degradation). Consequently, Fisher selection's advantage at pruning time (Section~\ref{sec:wap-selection-results}) does not carry over cleanly to robustness against \emph{further}, unplanned AP loss on top of that prune level.

RSSI noise (Table~\ref{tab:robustness}b), however, behaves differently. The three conditions track each other closely at low to moderate noise ($\sigma\leq0.05$, within about a percentage point of each other), yet they diverge at the highest noise level we tested ($\sigma=0.15$), where Fisher pruning shows the smallest relative degradation of the three (36.9\%), actually beating the full model (42.8\%) and mutual information pruning (44.1\%). Read together with the floor accuracy finding under pruning in Section~\ref{sec:degradation}, namely that MSCP reaches \emph{higher} floor accuracy than the full model, this offers a second, independent hint that a well chosen smaller AP subset can act as a light regularizer against corrupted input, not merely as a cost saver.

\section{Discussion and Limitation}
\label{sec:discussion}

\textbf{Why does Fisher selection lose at heavy pruning on UJIIndoorLoc?} We attribute this to our greedy algorithm, which chooses APs one at a time to maximize average log det gain, a good strategy in expectation. Nevertheless, it can end in a locally good, rather than globally optimal, order, especially when few APs remain and every choice matters. That the same score leads at every deep prune level on Tampere/TUT suggests the effect is tied to how much redundancy a site has left rather than to the criterion itself. As future work, therefore, we regard a Shapley value score~\cite{shapley1953value,cohen2005shapleyfeature,lundberg2017shap} (cooperative game theory, credit averaged over all orderings instead of one greedy path) and an Ollivier Ricci curvature score~\cite{ollivier2009ricci,topping2022curvature} (graph geometry, distinguishing "redundant" from "structurally critical" APs in a similarity graph) as candidate fixes.

\textbf{A per AP exponent and direct trilateration.} A further limitation lies in how the path loss exponent is parameterized. Each building receives its own exponent and every AP only a small deviation around it, which keeps training stable but cannot describe an AP whose propagation genuinely departs from its building average. The saturation analysis of Section~\ref{sec:geom-diag} makes this concrete, with roughly a third of APs pinned at the permitted boundary. We therefore plan to learn a fully free per AP exponent $\eta_j$. This matters for more than goodness of fit. Once every AP carries its own reference power, exponent, and coordinate, the physics decoder becomes individually invertible, so a measured RSSI vector yields per AP distance estimates and position can be recovered by classical trilateration over the learned geometry instead of by the discriminative head.

\textbf{Locating blind spots.} The recovered geometry should also be able to answer a question this paper does not, namely where to collect new fingerprint samples. The Fisher matrix of Section~\ref{sec:fisher-selection} is defined at every position, so its log determinant maps directly to a coverage surface over the building, and the regions where it is smallest are the regions that the retained access points constrain least. A natural design is an acquisition rule that combines this geometric term with the predictive uncertainty of the position head, obtained for instance by MC dropout~\cite{gal2016dropout}, and that is measured against spatial coverage rules such as k centre selection~\cite{sener2018coreset}. The design and evaluation of such a rule, together with the survey protocol it implies, is left to future work. We also plan to report proper run to run variance instead of point estimates for the geometry diagnostic (Section~\ref{sec:geom-diag}) and robustness (Section~\ref{sec:robustness}) suites.

All experiments ran on a single server GPU.

\section{Conclusion}
\label{sec:conclusion}

We treated interpretability as the design objective rather than a commentary added after the fact, presenting a physics informed WiFi localization model in which an effective AP geometry is recovered from RSSI and user positions, with no real AP location ever given. The interpretable configuration costs nothing (full model 7.07\,m vs.\ an ablated band of 7.21 to 7.29\,m, Section~\ref{sec:ablations}), the recovered geometry is faithful (randomized control correlation collapses from $-0.43$ to $-0.004$, Section~\ref{sec:geom-diag}), and it is useful. It supplies the AP coordinates that a classical D optimal design criterion requires and a fingerprint database never ships, so access points can be ranked by Fisher information with no surveyed position at all. The resulting geometry only selector is the equal of scores built directly from the labelled data on UJIIndoorLoc (Section~\ref{sec:wap-selection-results}), and the same substitution carries to two further independently collected sites, Tampere/TUT and UTSIndoorLoc, where it gives the lowest 3D error of any method on Tampere/TUT from 35\% pruned onward and retains the highest floor accuracy of any method on the 16 floor UTS building at the three deeper prune levels. Throughout, it leaves behind an inspectable artifact no statistical score produces. The model is also competitive at the underlying task, beating classic WKNN, KNN and MLP baselines by 26 to 36\% mean 3D error on UJIIndoorLoc.

The artifact has a clear operating range. Parity with the strongest classic method fails at 65\% pruned on UJIIndoorLoc, a third of the per AP exponents are pinned at their bound, and Fisher pruning is not uniformly the most robust choice under further AP loss. Section~\ref{sec:discussion} details the reasoning behind these limitations and the future work they motivate.

\bibliographystyle{IEEEtran}
\bibliography{refs}

\end{document}